\pdfoutput=1

\documentclass[11pt]{article}

\usepackage[]{ACL2023}

\usepackage{times}
\usepackage{latexsym}
\usepackage{float}
\usepackage{svg}
\usepackage{caption}
\usepackage{subcaption}
\usepackage{footmisc}
\usepackage{hyperref}

\usepackage[T1]{fontenc}

\usepackage[utf8]{inputenc}

\usepackage{microtype}

\usepackage{inconsolata}

\newcommand*\samethanks[1][\value{footnote}]{\footnotemark[#1]}

\title{Unsupervised Open-domain Keyphrase Generation}

\author{Lam Thanh Do$^{\spadesuit}$\thanks{\ \ {\footnotesize Work done while visiting Cazoodle Inc.}} $\quad$ Pritom Saha Akash$^{\heartsuit}$ $\quad$ Kevin Chen-Chuan Chang$^{\heartsuit}$\samethanks\\
  ${}^{\spadesuit}$Hanoi University of Science and Technology, Viet Nam \\
  ${}^{\heartsuit}$University of Illinois at Urbana-Champaign, USA \\
  \texttt{lam.dt183573@sis.hust.edu.vn} \\
  \texttt{\{pakash2, kcchang\}@illinois.edu}}

\usepackage[utf8]{inputenc}
\usepackage{amsmath}
\usepackage{amsfonts}
\usepackage{amssymb}
\usepackage{graphicx}
\usepackage{multirow}
\usepackage[normalem]{ulem}
\useunder{\uline}{\ul}{}

\usepackage{booktabs}
\usepackage{multirow}

\begin{document}
\maketitle
\begin{abstract}
In this work, we study the problem of unsupervised open-domain keyphrase generation, where the objective is a keyphrase generation model that can be built without using human-labeled data and can perform consistently across domains. To solve this problem, we propose a seq2seq model that consists of two modules, namely \textit{phraseness} and \textit{informativeness} module, both of which can be built in an unsupervised and open-domain fashion. The phraseness module generates phrases, while the informativeness module guides the generation towards those that represent the core concepts of the text. We thoroughly evaluate our proposed method using eight benchmark datasets from different domains. Results on in-domain datasets show that our approach achieves state-of-the-art results compared with existing unsupervised models, and overall narrows the gap between supervised and unsupervised methods down to about 16\%. Furthermore, we demonstrate that our model performs consistently across domains, as it overall surpasses the baselines on out-of-domain datasets.\footnote{Code and data will be available at \url{https://github.com/ForwardDataLab/UOKG}.}.
\end{abstract}

\section{Introduction}

\begin{figure*}[]
\begin{center}
\includegraphics[width=16cm]{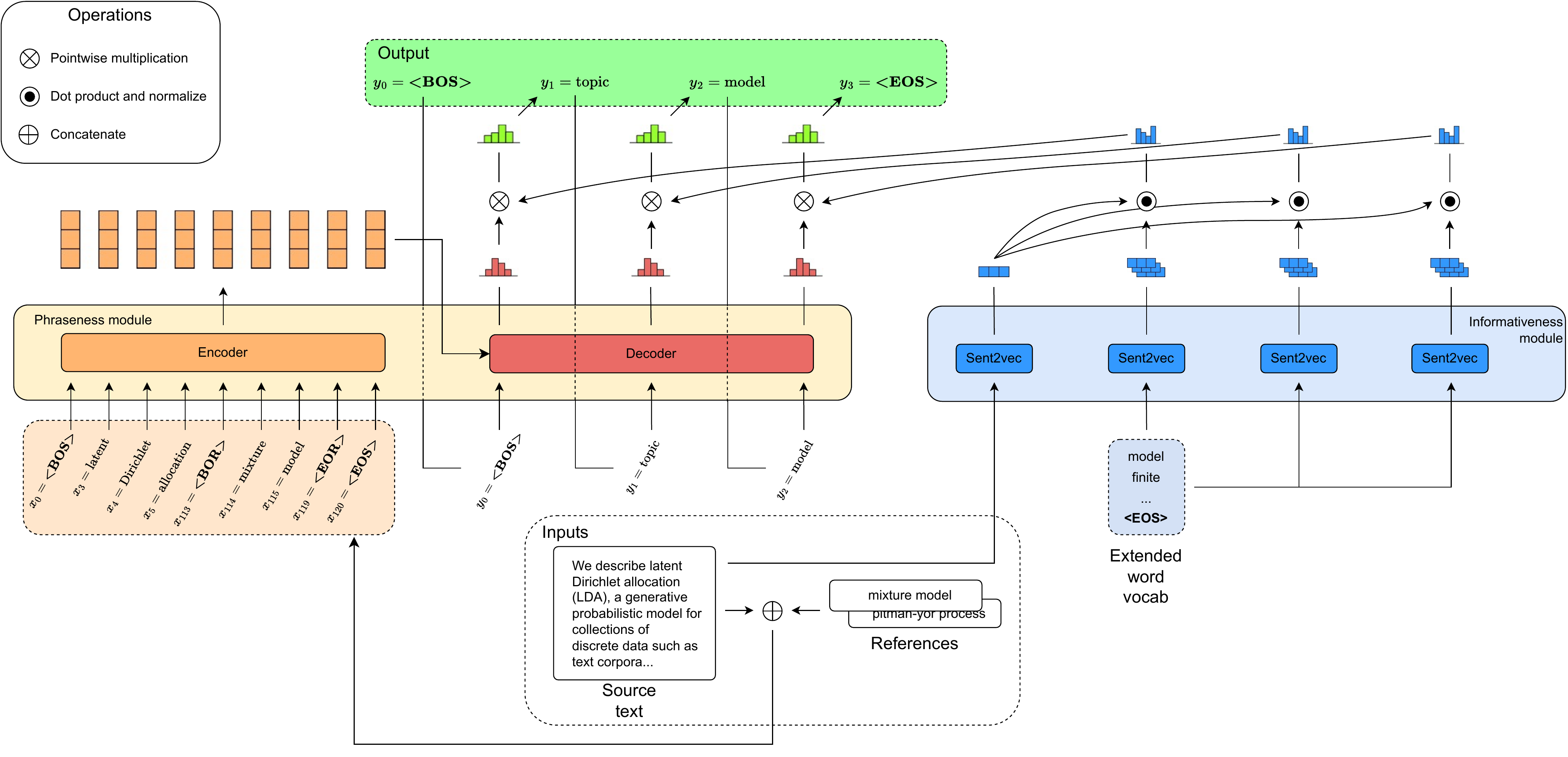}
\end{center}
\caption{Overview of our proposed model}
\label{fig:frameworkoverview}
\end{figure*}

Keyphrases are short word sequences that describe the core concepts of the text. The prediction of keyphrases for a text is a task that has received much attention recently. It is a crucial problem as its outputs can be useful for a variety of downstream tasks such as building digital libraries \cite{gutwin1999improving, witten2009build}, document summarization \cite{litvak-last-2008-graph}, document visualization \cite{chuang2012without} and so on.    

There are mainly two approaches to keyphrase prediction, namely \textit{keyphrase extraction} \cite{mihalcea-tarau-2004-textrank, florescu2017positionrank, bennani-smires-etal-2018-simple} and \textit{keyphrase generation} \cite{meng-etal-2017-deep, chen2019guided, yuan-etal-2020-one, Shen_Wang_Meng_Shang_2022}. Keyphrase extraction \textit{highlights} keyphrases that appear within the text. On the other hand, keyphrase generation \textit{generates} keyphrases based on the understanding of the given text and therefore allows predicting absent keyphrases alongside present ones \cite{meng-etal-2017-deep}. This ability has made keyphrase generation receive more attention than keyphrase extraction in recent years, as human also tend to use keyphrases that are absent from the text.

Most of the existing keyphrase generation models use manually labeled data for training \cite{meng-etal-2017-deep, chen-etal-2018-keyphrase, chen2019guided, yuan-etal-2020-one, ahmad-etal-2021-select}. However, obtaining labeled data is often the most expensive component of any machine learning model, and this is the same for keyphrase generation. Compared to labeled data, access to unlabeled data is easier and mostly available. For example, the arXiv dataset \cite{clement2019arxiv} containing metadata (e.g., title, abstract) of 1.7 million research articles is readily available on Kaggle. Therefore, it is more desirable to construct a keyphrase generation model in an unsupervised fashion. Furthermore, in practice, the model may have to encounter texts that come from various domains or even unseen ones. Therefore, another attractive property of a keyphrase generation model is the ability to handle open-domain documents.

Considering the above scenario, we propose a new problem called \textbf{Unsupervised Open-domain Keyphrase Generation}. Similar to every keyphrase generation methods, the model of our objective is given a text $\boldsymbol{x}$ as input, and as output, it generates a set of keyphrases $\{\boldsymbol{y}\}$. Both $\boldsymbol{x}$ and $\boldsymbol{y}$ are word sequences. Furthermore, the model of our objective should satisfy two requirements: 1) it can be built using only an unlabeled corpus, denoted as $D$; 2) it can effectively handle inputs from across domains.




This is a challenging task because we do not have access to labeled data from which to learn the patterns for keyphrases. Additionally, we also need our model to work across domains. This is difficult because there might exist different patterns for keyphrases for different domains. None of the existing work addresses these challenges. For instance, supervised keyphrase generation models \cite{meng-etal-2017-deep, chen-etal-2018-keyphrase, chen2019guided, yuan-etal-2020-one, ahmad-etal-2021-select} not only require manually labeled data for training but are also known to perform poorly when being moved out-of-domain. On the other hand, \cite{Shen_Wang_Meng_Shang_2022} propose AutoKeyGen, which uses pseudo-labeled data to train a seq2seq model in a weakly-supervised fashion, thereby removing the need for human annotation effort. However, similar to supervised models, the weakly-supervised approach taken by AutoKeyGen does not enable it to maintain performance in unseen domains.

Therefore, to solve our problem, we propose an unsupervised keyphrase generation model that can work across domains. The \textbf{key idea} is to modularize a seq2seq model into two modules. The motivation for modularizing is to decompose keyphrase generation into two simpler problems where each of which can be addressed in an unsupervised and open-domain setting. The first module, named the \textit{phraseness} module, is responsible for generating phrases, while the second module, named the \textit{informativeness} module, guides the generation toward the phrases that represent the most crucial concepts of the text.  



The phraseness module is a retrieval-augmented seq2seq model, where the retriever assists the seq2seq component in generating absent phrases alongside present ones. This module can be built in an unsupervised fashion because it leverages noun phrases to index the retriever and to train the seq2seq model, which can easily be obtained using open-sourced software libraries such as NLTK \cite{bird2009natural}, and therefore does not require human annotation effort. Furthermore, the phraseness module can also be built in an open-domain fashion, thanks to 1) the part-of-speech information incorporated into the seq2seq model, which allows copying words to form grammatically correct noun phrases regardless of domains; 2) the fact that the retriever can be further indexed with domain-relevant information, to provide reliable references.



The informativeness module is another seq2seq model, where a phrase is likely to be generated if it contains words that are informative to the given text. Inspired by embedding-based unsupervised keyphrase extraction (UKE) methods, we quantify informativeness of a word and a text based on their closeness in meaning, which is measured via the similarity between their embeddings. We choose this method of evaluating informativeness over other UKE methods (e.g. graph-based, statistics based) since it supports not only present phrases, but also absent ones.  Similar to the phraseness module, the informativeness module can also be built in an unsupervised and open-domain fashion. This is obtained by using a domain-general, unsupervised text embedding model (e.g. Sent2Vec \cite{pagliardini-etal-2018-unsupervised}).

We summarize the contributions of our paper. \textbf{Firstly}, we propose a new problem called \textit{unsupervised open-domain keyphrase generation}. \textbf{Secondly}, we design a model for solving the problem. Our proposed model is a seq2seq model that consists of two modules, one is responsible for generating phrases and the other guides the generation towards the phrases that represent the core concepts of the text. \textbf{Finally}, we conduct extensive experiments on multiple datasets across domains to demonstrate the effectiveness of our model as we contrast it against multiple strong baselines.

\section{Proposed method}

Figure \ref{fig:frameworkoverview} illustrates our proposed framework. We propose a seq2seq model that consists of two modules, namely \textit{phraseness} and i\textit{nformativeness} module. We adopt the two terms \textit{phraseness} and \textit{informativeness} from \cite{tomokiyo-hurst-2003-language}, to describe the desirable criteria a keyphrase should satisfy. \textit{Phraseness} refers to the degree to which a word sequence is considered a phrase, and \textit{informativeness} refers to how well the phrase illustrates the core concepts of the text. Each of the two modules guarantees a criterion mentioned above. In particular, the phraseness module generates (present and absent) phrases, while the informativeness module guides the generation toward phrases that describe the core concepts of the text. In the following sections, we will describe in detail the two modules, as well as how they are combined to generate keyphrases.

\begin{figure}[h]
\begin{center}
\includegraphics[width=8cm]{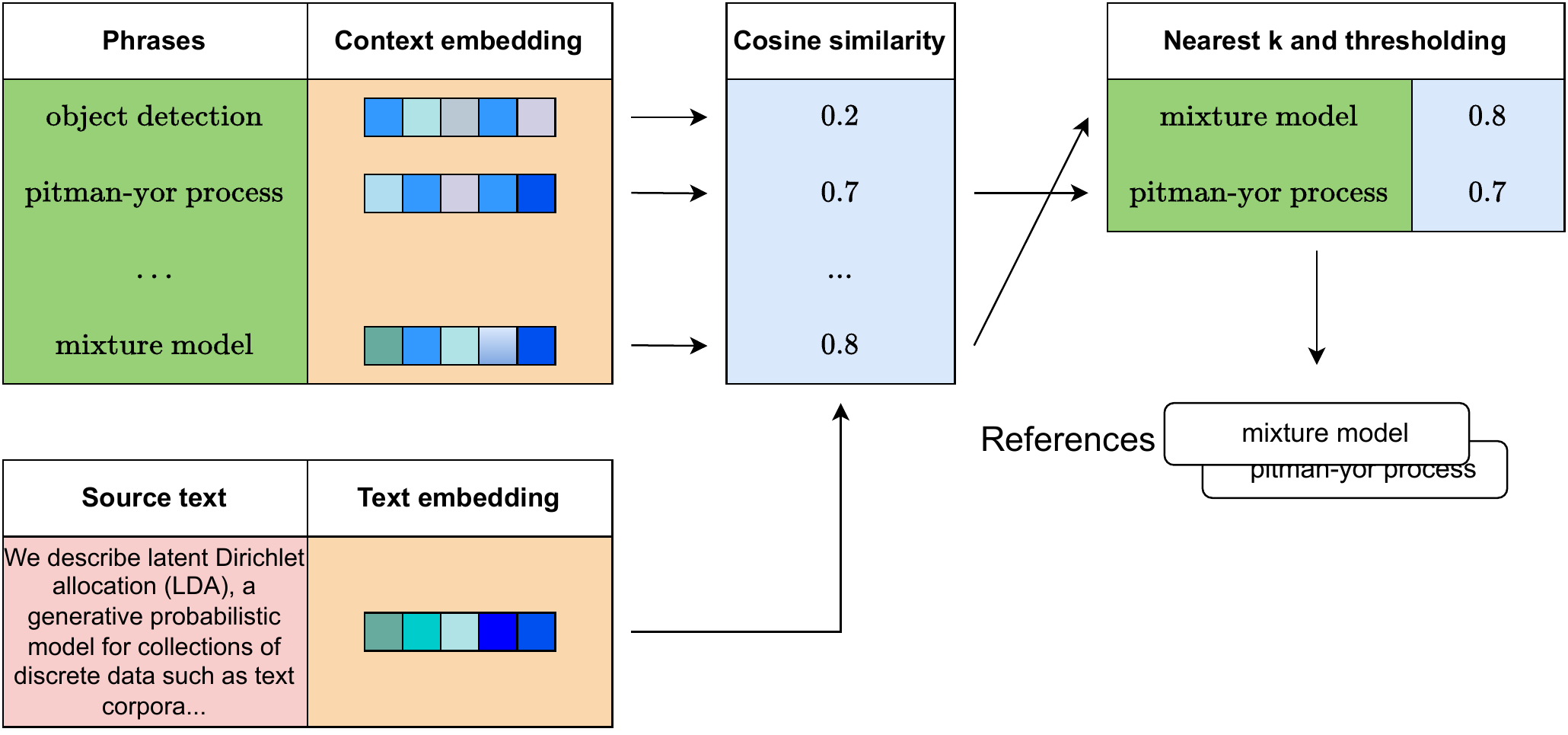}
\end{center}
\caption{Illustration of the retrieval of references.}
\label{fig:retrieve_references}
\end{figure}

\subsection{Phraseness module}

In order to generate keyphrases, it is crucial to know how to first generate phrases. We emphasize the difference between a keyphrase and a phrase - the former needs to be informative to the given text, while the latter does not. It has been shown that keyphrases mostly take the form of noun phrases \cite{chuang2012without}. Also, recent work on keyphrase generation has shown that absent keyphrases can often be retrieved from other texts \cite{ye-etal-2021-heterogeneous}, suggesting that absent phrases can be found similarly. Therefore, a simple solution to obtaining phrases is to extract noun phrases as present phrases and retrieve related noun phrases as absent ones. 

\begin{figure*}[h]
\centering
     \begin{subfigure}[b]{\columnwidth}
         \centering
         \includegraphics[width=\textwidth]{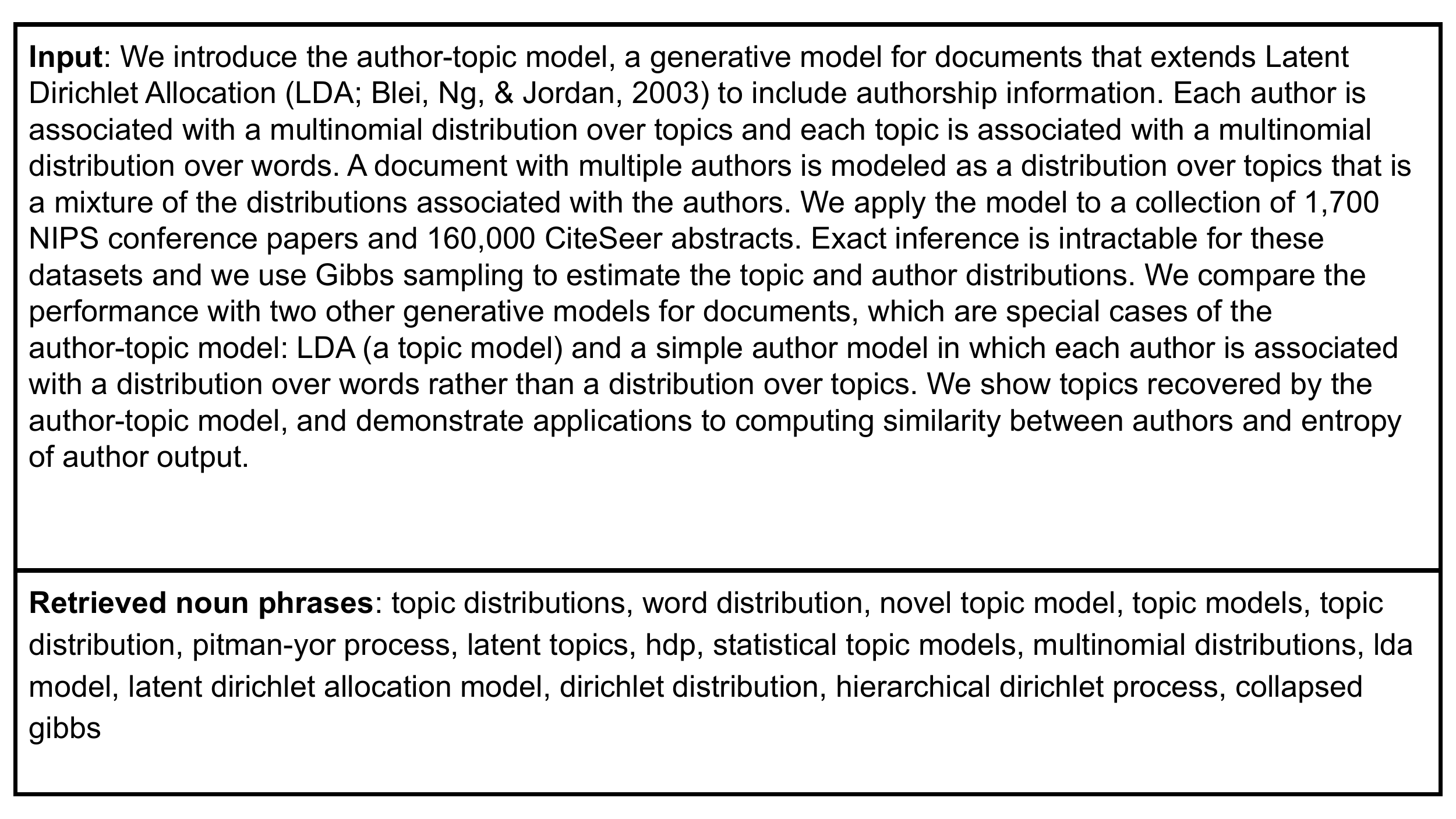}
         \caption{Example of using retrieval to predict \\ absent phrases}
         \label{fig:example_phraseness}
     \end{subfigure}
     \hfill
     \begin{subfigure}[b]{\columnwidth}
         \centering
         \includegraphics[width=\textwidth]{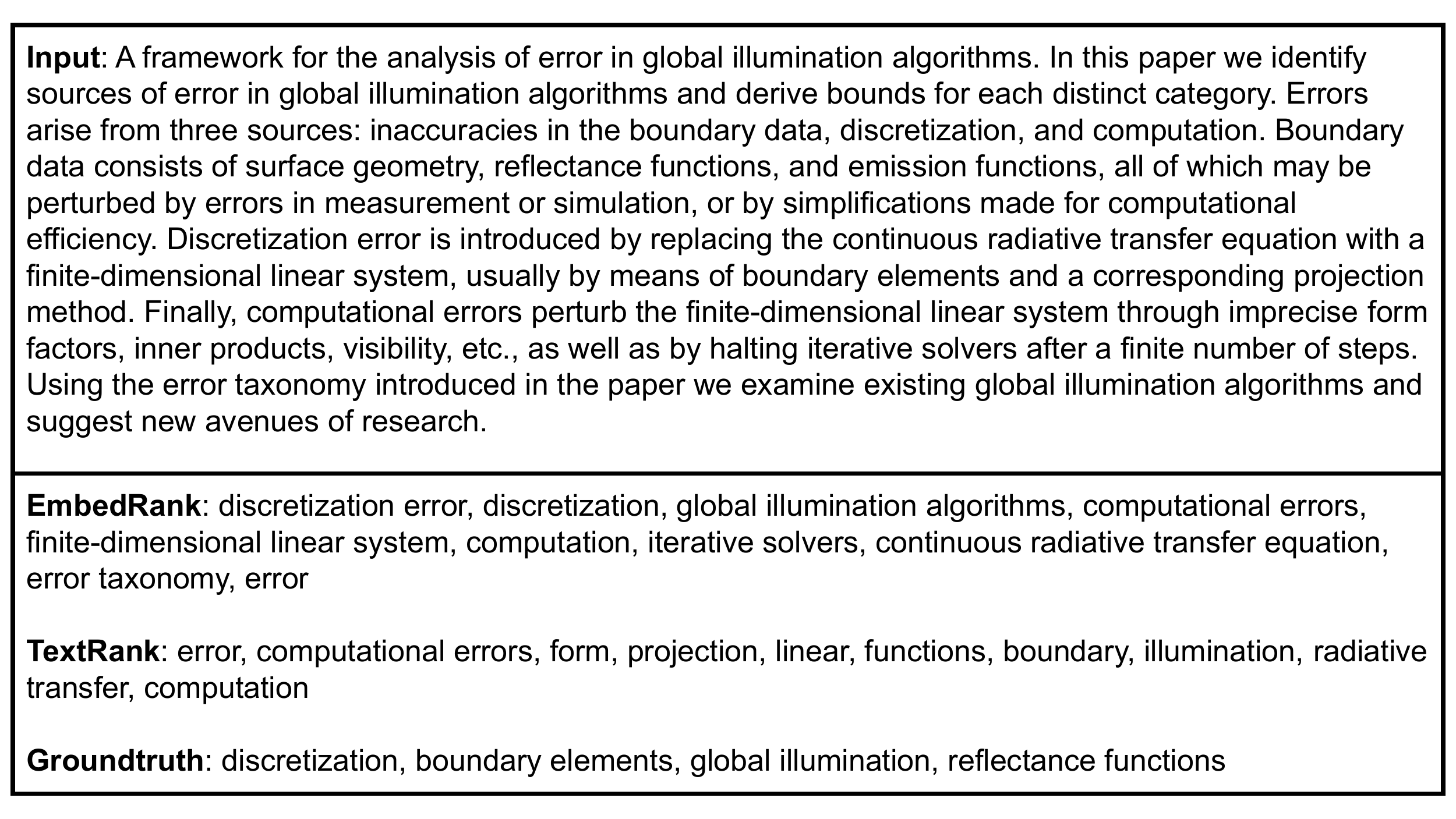}
         \caption{Example of EmbedRank and TextRank's prediction,\\ along with the groundtruth keyphrases}
         \label{fig:example_combining}
     \end{subfigure}
     \caption{Examples}
\end{figure*}

However, this simple solution may not be optimal. Since the retrieved phrases are originally used in other texts, they may not be suitable to describe the concepts of the given text. We demonstrate this limitation using the example in Figure \ref{fig:example_phraseness}. In this example, the absent phrases obtained via retrieval describe concepts related to ``topic modeling''. However, our desired outputs need to also describe concepts related to ``author modeling''.

The above problem could be mitigated if we also consider the given text alongside the retrieved noun phrases. In the example above, relevant phrases such as ``author topic distributions'' can be generated by combining ``author'', which is from the given text, and ``topic distributions'', which is one of the retrieved phrases. With this in mind, we employ a \textit{retrieval-augmented seq2seq model} as the phraseness module. First, a set of related but absent noun phrases is retrieved, which we will now refer to as \textit{references}. Then, a seq2seq model generates noun phrases based on both the text and the references.


\subsubsection{Retriever}

Figure \ref{fig:retrieve_references} describes the retrieval of references given a text. To obtain references for the input, we leverage existing noun phrases observed in other documents. We assume that a noun phrase is related to a text if it occurs in contexts similar to that text. With this in mind, we collect noun phrases from documents in the unlabeled corpus $D$ to form a phrase bank $B$. We index each noun phrase $\boldsymbol{z} \in B$ with a \textit{context embedding}, denoted as $c_{\boldsymbol{z}}$, which is obtained by averaging the embeddings of the documents in which $\boldsymbol{z}$ appears in. We obtain the embeddings of texts by using Sent2Vec \cite{pagliardini-etal-2018-unsupervised}, an unsupervised sentence embedding model. To retrieve references for a text $\boldsymbol{x}$, we first use Sent2Vec to compute its embedding, denoted as $v_{\boldsymbol{x}}$, and then retrieve the top-$k$ phrases $z$ based on the following retrieval score $R_{\boldsymbol{x}}(\boldsymbol{z}) = \cos(c_z, v_{\boldsymbol{x}})$. Furthermore, in order to prevent retrieving unreliable references, we filter those whose retrieval scores are smaller than a threshold $\tau$. We denote the set of references for $\boldsymbol{x}$ as $Z_{\boldsymbol{x}}$.

As mentioned above, we can further index the retriever with other corpora, denoted as $D^\prime$, from different domains. To do this, all we need to do is to update the phrase bank $B$ with new phrases from $D^\prime$ and update the context embeddings of every phrase that occur in both $D$ and $D^\prime$.

\subsubsection{Seq2Seq model}

\textbf{Input representation.} The seq2seq model takes as inputs not only the source text $\boldsymbol{x}$ but also its references $Z_{\boldsymbol{x}}$, to generate phrases. The text and its references are combined into a single input $\boldsymbol{\Tilde{x}}$, defined as

\resizebox{0.9\linewidth}{!}{
\begin{minipage}{\linewidth}
\vspace{-4.5mm}
\begin{align}
    \boldsymbol{\Tilde{x}} = \text{{\fontfamily{qcr}\selectfont [BOS]}} \ \boldsymbol{x} \ \text{{\fontfamily{qcr}\selectfont [BOR]}} \ Z_{\boldsymbol{x}} \ \text{{\fontfamily{qcr}\selectfont [EOR]}} \ \text{{\fontfamily{qcr}\selectfont [EOS]}} 
\end{align}
  \end{minipage}
}

where $\text{{\fontfamily{qcr}\selectfont [BOS]}}$ and $\text{{\fontfamily{qcr}\selectfont [EOS]}}$ are respectively the beginning and end of sentence token. The two tokens $\text{{\fontfamily{qcr}\selectfont [BOR]}}$ and $\text{{\fontfamily{qcr}\selectfont [EOR]}}$ signals the start and end of the reference block. In addition, the references are separated by a $\text{{\fontfamily{qcr}\selectfont [SEP]}}$ token.

\textbf{Model architecture.} We employ Transformer \cite{vaswani2017attention} with copy mechanism \cite{gu-etal-2016-incorporating, see-etal-2017-get} as the architecture of our seq2seq model. First, the encoder receives the word embeddings of the tokens in $\boldsymbol{\Tilde{x}}$, producing a sequence of encoder hidden states $\boldsymbol{h} = \{h_i\}_{i=1}^{|\boldsymbol{\Tilde{x}}|}$. The decoder takes the embeddings of the previously generated words $\boldsymbol{y}_{<t}$ and the encoder hidden states, outputting the decoder hidden state $s_t$. For each input, we build an extended vocabulary $V_{\boldsymbol{\Tilde{x}}}$, which is the union of the decoder's vocabulary $V$ and the words in the augmented input $\boldsymbol{\Tilde{x}}$. Finally, we compute the phraseness probability of predicting a word from $V_{\boldsymbol{\Tilde{x}}}$ as $P_{\text{pn}}(y_t|\boldsymbol{y}_{<t}, \boldsymbol{\Tilde{x}}) = p_{\text{gen}} P^V_{\text{pn}}(y_t|\boldsymbol{y}_{<t}, \boldsymbol{\Tilde{x}}) + (1 - p_{\text{gen}}) P^C_{\text{pn}}(y_t|\boldsymbol{y}_{<t}, \boldsymbol{\Tilde{x}})$. Here, $P^V_{\text{pn}} (y_t|\boldsymbol{y}_{<t}, \boldsymbol{\Tilde{x}}) = \text{{\fontfamily{qcr}\selectfont softmax}}(W^V s_t)$ is the distribution over the word vocabulary $V$, $p_{\text{gen}}=\text{{\fontfamily{qcr}\selectfont sigmoid}}(W^{\text{g}}_s s_t + W^{\text{g}}_y y_{t-1})$ is the soft switch between generating and copy. All the $W$ terms are trainable parameters, and we omit the bias terms for less cluttered notation.

We incorporate part-of-speech information to copy words from $\boldsymbol{\Tilde{x}}$. More formally, $P^C_{\text{pn}}(y_t=w|\boldsymbol{y}_{<t}, \boldsymbol{\Tilde{x}}) = \sum_{\Tilde{x_i}=w} a_i^t$, where $a_i^t = \text{{\fontfamily{qcr}\selectfont softmax}}(e_i^t)$, and $e_i^t = \text{{\fontfamily{qcr}\selectfont FF}}_h(\tilde{h_i}^T) \text{{\fontfamily{qcr}\selectfont FF}}_s (\tilde{s_t})$. Here, $\tilde{h_i} = \text{{\fontfamily{qcr}\selectfont concat}} (h_i, l_{\Tilde{x}_i})$ is the encoder hidden state of $\Tilde{x}_i$ enhanced with its part-of-speech embedding $l_{\Tilde{x}_i}$. Similarly, $\tilde{s_t}=\text{{\fontfamily{qcr}\selectfont concat}} (s_t, l_{y_{t}})$ is the decoder hidden state enhanced with the part-of-speech embedding of the previously generated word. $\text{{\fontfamily{qcr}\selectfont FF}}_h$ and $\text{{\fontfamily{qcr}\selectfont FF}}_s$ denotes the feedforward neural networks, whose purposes are to help project $\tilde{h_i}$ and $\tilde{s_t}$ into the same semantic space.

\textbf{Model training.} For every document $\boldsymbol{x_i} \in D$, we maximize $\log P_{\text{pn}}(\boldsymbol{y}=\boldsymbol{z}|\boldsymbol{\Tilde{x}})$, where

\resizebox{0.9\linewidth}{!}{
\begin{minipage}{\linewidth}
\vspace{-4.5mm}
\begin{align}
    P_{\text{pn}}(\boldsymbol{y}=\boldsymbol{z}|\boldsymbol{\Tilde{x}})=\prod_{t=1}^{T} P_{\text{pn}}(y_t=z_t|\boldsymbol{y}_{<t}, \boldsymbol{\Tilde{x}})
\end{align}
  \end{minipage}
}

for the phrases $\boldsymbol{z}=\{z_t\}$, which include the present noun phrases and the references. To encourage the model to generate absent phrases instead of just copying it from the references, we randomly mask some references and train the model to generate them.

\subsection{Informativeness module}

Knowing to generate phrases is not sufficient to obtain keyphrases. It is also important to guide the generation towards the phrases that are informative to the input. Previous work on unsupervised keyphrase extraction offer multiple classes of methods, namely graph-based, statistics-based and embedding-based, for evaluating informativeness of phrases (for more details, see Section \ref{sec:related_word}). Graph-based and statistics-based methods are not suitable in our setting. These methods utilize only in-text information and therefore cannot determine the informativeness of absent phrases. On the other hand, embedding-based methods evaluate informativeness of a phrase based on its closeness in meaning with the input text. As a result, these methods can support both present and absent phrases. We therefore adopt the idea of embedding-based methods in building our informativeness module.

Let us define $S(\boldsymbol{a}, \boldsymbol{b})=\max (0, v_{\boldsymbol{a}}^T v_{\boldsymbol{b}})$ as the similarity score between two pieces of text, where $v_{\boldsymbol{a}}$, $v_{\boldsymbol{b}}$ are embeddings obtained using Sent2Vec. Using this score, we define the informativeness distribution $P_{\text{in}}(\boldsymbol{y}|\boldsymbol{x})$, by decomposing it into conditional distributions of each word given the previous context. More formally, $P_{\text{in}}(\boldsymbol{y}|\boldsymbol{x}) = \prod_{t=1}^{T} P_{\text{in}}(y_t|\boldsymbol{y}_{<t}, \boldsymbol{x})$, where

\resizebox{0.9\linewidth}{!}{
\begin{minipage}{\linewidth}
\vspace{-4.5mm}
\begin{align}
     P_{\text{in}}(y_t=w|\boldsymbol{y}_{<t}, \boldsymbol{x}) \propto \begin{cases}
    S(w,\boldsymbol{x}), & \text{if $w \neq \text{{\fontfamily{qcr}\selectfont [EOS]}}$} \\
    S(\boldsymbol{y}_{<t}, \boldsymbol{x}), & \text{otherwise}
     \end{cases}
    \end{align}
  \end{minipage}
}


The probability $P_{\text{in}}(y_t=w|\boldsymbol{y}_{<t}, \boldsymbol{x})$ is normalized over the extended word vocabulary $V_{\boldsymbol{\Tilde{x}}}$, which is the same one used by the phraseness module. Intuitively, a word has high probability of being generated if that word has close meaning to the text. The {\fontfamily{qcr}\selectfont [EOS]} token is likely to be generated if the currently generated phrase $\boldsymbol{y}_{<t}$ already form an informative phrase.

\subsection{Combining phraseness and informativeness}
Generating keyphrases require us to enforce both phraseness and informativeness on the output sequence. A simple solution is to adopt the approaches taken by existing unsupervised keyphrase extraction methods, which enforce the two criteria sequentially. In particular, they either 1) form phrases first, then choose those that are most informative as keyphrases; or 2) choose informative words first, then form keyphrases using these words. However, both approaches may not be optimal. The first approach may include uninformative words in the prediction, while the second rigidly assume that a keyphrase should only contain keywords. We illustrate the limitation of these approaches using an example, shown in Figure \ref{fig:example_combining}. Here, we show the predictions of EmbedRank \cite{bennani-smires-etal-2018-simple}, which takes approach 1) and TextRank \cite{mihalcea-tarau-2004-textrank}, which takes approach 2). Both of them fail to predict the golden keyphrase ``global illumination''. EmbedRank redundantly include the word ``algorithms'', while TextRank only outputs ``illumination'', as ``global'' is not predicted as a keyword.

This problem could be alleviated if both phraseness and informativeness is considered when forming the keyphrase. In the example above, the word ``algorithms'' should be excluded, since it neither contributes to the informativeness of the phrase, nor it is required to make the phrase understandable. On the other hand, the word ``global'' may not be among the most informative words to the text, however, this word is essential as excluding it results in a phrase with a different concept.

In light of this, we propose to generate keyphrases, one word at a time, where each word is generated if it is predicted by both the phraseness and informativeness module. To this end, we propose to combine the two modules in a product-of-experts fashion \cite{hinton2002training}. In particular, the conditional distribution of a keyphrase given a text is defined as follows

\resizebox{0.9\linewidth}{!}{
\begin{minipage}{\linewidth}
\vspace{-3.5mm}
\begin{align}
\begin{split}
P_{\text{kp}}(\boldsymbol{y}|\boldsymbol{x}) &\propto P_{\text{pn}}(\boldsymbol{y}|\boldsymbol{\Tilde{x}}) ^ {\lambda} \cdot P_{\text{in}}(\boldsymbol{y}|\boldsymbol{x}) \\
&\propto \prod_{t=1}^{T} P_{\text{pn}}(y_t|\boldsymbol{y}_{<t}, \boldsymbol{\Tilde{x}})^{\lambda} \cdot P_{\text{in}}(y_t|\boldsymbol{y}_{<t}, \boldsymbol{x})
\end{split}
\end{align}
\end{minipage}
}

where $\lambda$ is a hyperparameter for balancing the two modules. 

The idea of combining two language models using the product-of-experts has previously been studied for the task of unsupervised abstractive summarization \cite{zhou-rush-2019-simple}. To the best of our knowledge, we are the first to use this idea in unsupervised keyphrase generation. In the above paragraphs, we also discussed why it is a suitable choice.

\subsection{Keyphrase decoding}
To decode keyphrases, we employ beam search to search for keyphrases based on $s(\boldsymbol{y}) = - \log P_{\text{kp}}(\boldsymbol{y}|\boldsymbol{x})$. As beam search tend to favor shorter keyphrases, we employ the length normalization strategy similarly to that described in \cite{sun2019divgraphpointer}, which is to divide $s(\boldsymbol{y})$ by $|\boldsymbol{y}| + \alpha$, where $\alpha$ is a length penalty factor.

It has been shown in previous work that positional information is useful for the prediction of present keyphrases \cite{florescu-caragea-2017-positionrank, gallina2020large}. Therefore, it is desirable to incorporate this feature into our model. Furthermore, we found that the model tends to generate absent keyphrases that are entirely new. This behavior may not be desirable for downstream tasks such as document retrieval, where we need to associate documents with common keyphrases. Based on the above discussion, we propose to rerank the beam search results using the following score

\resizebox{0.9\linewidth}{!}{
\begin{minipage}{\linewidth}
\vspace{-4.5mm}
\begin{align}
    \hat{s}(\boldsymbol{y}) = \frac{s(\boldsymbol{y})}{|\boldsymbol{y}| + \alpha} \times b(\boldsymbol{y})
\end{align}
\end{minipage}
}


\resizebox{0.9\linewidth}{!}{
\begin{minipage}{\linewidth}
\vspace{-4.5mm}
\begin{align}
     \label{eq:adjustment_weight}
b(\boldsymbol{y})=\begin{cases}
			\beta, & \text{if $\boldsymbol{y}$ is absent and $\boldsymbol{y} \in B$}\\
   1, & \text{if $\boldsymbol{y}$ is absent and $\boldsymbol{y} \not \in B$}\\
            \frac{\log_2 (1 + \mathcal{P}_{\boldsymbol{x}}(\boldsymbol{y}))}{\log_2 (1+\mathcal{P}_{\boldsymbol{x}}(\boldsymbol{y})) + 1}, & \text{if $\boldsymbol{y}$ is present}
		 \end{cases}
    \end{align}
  \end{minipage}
}

where $b(\boldsymbol{y})$ is an adjustment weight, $\beta$ is a hyperparameter for adjusting the scores of absent phrases that exist in the phrase bank $B$ ($\beta < 1$ indicates that we favor $\boldsymbol{y} \in B$), and $\mathcal{P}_{\boldsymbol{x}}(\boldsymbol{y})$ is the word offset position of the phrase $\boldsymbol{y}$ in the text $\boldsymbol{x}$. Intuitively, $b(\boldsymbol{y})$ favors present keyphrases that appear earlier in the text, and absent keyphrases that exist in the phrase bank $B$.

\begin{table}[]
\resizebox{\columnwidth}{!}{%
\begin{tabular}{|l|l|l|l|l|l|l|}
\hline
Dataset name  & Language & Type                                                        & valid/test docs & \#kps/doc & \%absent & \%overlap \\ \hline
SemEval  & English & Scientific & 144/100 & 15.4 & 58.2 & 38.8 \\
Inspec   & English & Scientific & 1500/500 & 9.7  & 22.7 & 40   \\
NUS      & English & Scientific & 50/161  & 11.6 & 53.1 & 53.3 \\
Krapivin & English & Scientific & 1844/460 & 5.3  & 50.5 & 53.5 \\
StackExchange & English  & \begin{tabular}[c]{@{}l@{}}Technical \\ Question\end{tabular} & 16000/16000              & 2.7              & 48.9     & 45.1      \\
DUC-2001 & English & News       & 50/268  & 8.1  & 2.7  & 14.7 \\
KPTimes  & English & News       & 10000/20000 & 5  & 46.2 & 7.5   \\
OpenKP   & English & News       & 6616/6614 & 2.2  & 10.9 & 12.7 \\ \hline
\end{tabular}%
}
\caption{Statistics of testing datasets.}
\label{tab:data_statistics}
\end{table}

\begin{table*}[]
\resizebox{\textwidth}{!}{%
\begin{tabular}{|lllllllllllll|}
\hline
\multicolumn{13}{|c|}{\textbf{Present keyphrase generation}} \\ \hline
\multicolumn{1}{|l|}{\multirow{2}{*}{}} &
  \multicolumn{2}{c|}{SemEval} &
  \multicolumn{2}{c|}{Inspec} &
  \multicolumn{2}{c|}{NUS} &
  \multicolumn{2}{c|}{Krapivin} &
  \multicolumn{2}{c|}{StackExchange} &
  \multicolumn{2}{c|}{Average} \\ \cline{2-13} 
\multicolumn{1}{|l|}{} &
  F1@3 &
  \multicolumn{1}{l|}{F1@5} &
  F1@3 &
  \multicolumn{1}{l|}{F1@5} &
  F1@3 &
  \multicolumn{1}{l|}{F1@5} &
  F1@3 &
  \multicolumn{1}{l|}{F1@5} &
  F1@3 &
  \multicolumn{1}{l|}{F1@5} &
  F1@3 &
  F1@5 \\ \hline
\multicolumn{1}{|l|}{TF-IDF} &
  19 &
  \multicolumn{1}{l|}{\textbf{23.9}} &
  18.7 &
  \multicolumn{1}{l|}{24.8} &
  22.7 &
  \multicolumn{1}{l|}{{\ul 25.9}} &
  15.9 &
  \multicolumn{1}{l|}{15.7} &
  {\ul 18.8} &
  \multicolumn{1}{l|}{{\ul 16.5}} &
  19 &
  {\ul 21.4} \\
\multicolumn{1}{|l|}{TextRank} &
  13.8 &
  \multicolumn{1}{l|}{17.2} &
  17.7 &
  \multicolumn{1}{l|}{25} &
  14.9 &
  \multicolumn{1}{l|}{18.9} &
  11.1 &
  \multicolumn{1}{l|}{13.3} &
  8.5 &
  \multicolumn{1}{l|}{8.4} &
  13.2 &
  16.6 \\
\multicolumn{1}{|l|}{MultipartiteRank} &
  18.9 &
  \multicolumn{1}{l|}{21.4} &
  23.1 &
  \multicolumn{1}{l|}{26.5} &
  22.7 &
  \multicolumn{1}{l|}{24.9} &
  19.3 &
  \multicolumn{1}{l|}{{18.5}} &
  13.9 &
  \multicolumn{1}{l|}{13.6} &
  {\ul 19.6} &
  21 \\
\multicolumn{1}{|l|}{EmbedRank} &
  17.9 &
  \multicolumn{1}{l|}{21.2} &
  \textbf{26.2} &
  \multicolumn{1}{l|}{\textbf{32.6}} &
  17.5 &
  \multicolumn{1}{l|}{20.8} &
  13.5 &
  \multicolumn{1}{l|}{15.2} &
  11.8 &
  \multicolumn{1}{l|}{12.6} &
  17.4 &
  20.5 \\
\multicolumn{1}{|l|}{Global-Local Rank} &
  \textbf{20.4} &
  \multicolumn{1}{l|}{{\ul 23.6}} &
  {\ul 24.5} &
  \multicolumn{1}{l|}{{\ul 30.6}} &
  22.4 &
  \multicolumn{1}{l|}{23.7} &
  15 &
  \multicolumn{1}{l|}{15.2} &
  10.2 &
  \multicolumn{1}{l|}{9.8} &
  18.5 &
  20.6 \\ \hline
\multicolumn{1}{|l|}{AutoKeyGen} &
  16.6$_4$ &
  \multicolumn{1}{l|}{22.1$_4$} &
  19.4$_2$ &
  \multicolumn{1}{l|}{23.1$_3$} &
  {\ul 23.2$_4$} &
  \multicolumn{1}{l|}{{25.7$_3$}} &
  {\ul 19.5$_7$} &
  \multicolumn{1}{l|}{\ul 20.6$_5$} &
  14$_8$ &
  \multicolumn{1}{l|}{14.9$_6$} &
  18.5$_3$ &
  21.3$_2$ \\
\multicolumn{1}{|l|}{\textbf{Ours}} &
  {\ul 19.1$_8$} &
  \multicolumn{1}{l|}{22.2$_9$} &
  19.8$_8$ &
  \multicolumn{1}{l|}{23.3$_11$} &
  \textbf{26.4$_7$} &
  \multicolumn{1}{l|}{\textbf{27.8$_4$}} &
  \textbf{22.2$_{8}$} &
  \multicolumn{1}{l|}{\textbf{21.4$_{9}$}} &
  \textbf{27.2$_1$} &
  \multicolumn{1}{l|}{\textbf{25.1$_2$}} &
  \textbf{23$_4$} &
  \textbf{24$_4$} \\ \hline
\multicolumn{1}{|l|}{Supervised - CopyRNN} &
  26.1$_4$ &
  \multicolumn{1}{l|}{29.7$_5$} &
  19.1$_5$ &
  \multicolumn{1}{l|}{22.8$_5$} &
  35.5$_{13}$ &
  \multicolumn{1}{l|}{37.9$_4$} &
  30.3$_9$ &
  \multicolumn{1}{l|}{30.1$_6$} &
  24.1$_{6}$ &
  \multicolumn{1}{l|}{22.4$_5$} &
  27$_6$ &
  28.6$_3$ \\ \hline
\multicolumn{13}{|c|}{\textbf{Absent keyphrase generation}} \\ \hline
\multicolumn{1}{|l|}{\multirow{2}{*}{}} &
  \multicolumn{2}{c|}{SemEval} &
  \multicolumn{2}{c|}{Inspec} &
  \multicolumn{2}{c|}{NUS} &
  \multicolumn{2}{c|}{Krapivin} &
  \multicolumn{2}{c|}{StackExchange} &
  \multicolumn{2}{c|}{Average} \\ \cline{2-13} 
\multicolumn{1}{|l|}{} &
  R@5 &
  \multicolumn{1}{l|}{R@10} &
  R@5 &
  \multicolumn{1}{l|}{R@10} &
  R@5 &
  \multicolumn{1}{l|}{R@10} &
  R@5 &
  \multicolumn{1}{l|}{R@10} &
  R@5 &
  \multicolumn{1}{l|}{R@10} &
  R@5 &
  R@10 \\ \hline
\multicolumn{1}{|l|}{UKE methods} &
  0 &
  \multicolumn{1}{l|}{0} &
  0 &
  \multicolumn{1}{l|}{0} &
  0 &
  \multicolumn{1}{l|}{0} &
  0 &
  \multicolumn{1}{l|}{0} &
  0 &
  \multicolumn{1}{l|}{0} &
  0 &
  0 \\ \hline
\multicolumn{1}{|l|}{AutoKeyGen} &
  0.7$_2$ &
  \multicolumn{1}{l|}{1.2$_3$} &
  1.7$_2$ &
  \multicolumn{1}{l|}{2.8$_4$} &
  1$_2$ &
  \multicolumn{1}{l|}{1.9$_5$} &
  2.4$_2$ &
  \multicolumn{1}{l|}{3.8$_5$} &
  1.2$_2$ &
  \multicolumn{1}{l|}{1.9$_1$} &
  1.4$_1$ &
  2.3$_2$ \\
\multicolumn{1}{|l|}{\textbf{Ours}} &
  \textbf{1.4$_2$} &
  \multicolumn{1}{l|}{\textbf{2.3$_4$}} &
  \textbf{2.1$_2$} &
  \multicolumn{1}{l|}{\textbf{3$_2$}} &
  \textbf{1.8$_8$} &
  \multicolumn{1}{l|}{\textbf{3.1$_5$}} &
  \textbf{4.5$_5$} &
  \multicolumn{1}{l|}{\textbf{7$_2$}} &
  \textbf{4.6$_1$} &
  \multicolumn{1}{l|}{\textbf{6.3$_2$}} &
  \textbf{2.9$_2$} &
  \textbf{4.3$_2$} \\ \hline
\multicolumn{1}{|l|}{Supervised - CopyRNN} &
  2.1$_3$ &
  \multicolumn{1}{l|}{2.7$_3$} &
  3.7$_3$ &
  \multicolumn{1}{l|}{5.3$_3$} &
  4.4$_4$ &
  \multicolumn{1}{l|}{6.4$_7$} &
  7.9$_2$ &
  \multicolumn{1}{l|}{10.7$_7$} &
  2.3$_2$ &
  \multicolumn{1}{l|}{3.5$_3$} &
  4.1$_1$ &
  5.7$_3$ \\ \hline
\end{tabular}%
}
\caption{Keyphrase generation performance for in-domain datasets. The best and second-best results for each category are bold and highlighted, respectively. For AutoKeyGen, CopyRNN and our model, we run experiments five times with different random seeds and report the average. The subscript denotes the corresponding standard deviation (e.g. 26.4$_7$ indicates 26.4$\pm$ 0.7). We report both F1 and Recall in percentage points.}
\label{tab:in_domain_performance}
\end{table*}

\section{Experiments}
\subsection{Datasets}
We use the documents from the training set of KP20K \cite{meng-etal-2017-deep} to train our model and to index the retriever in the training phase. It contains the abstracts and titles of 514k scientific articles. In the testing phase, we utilize 8 datasets, namely SemEval \cite{kim2013automatic}, Inspec \cite{hulth2003improved}, NUS \cite{nguyen2007keyphrase}, Krapivin \cite{krapivin2009large}, DUC-2001 \cite{wan2008single}, OpenKP \cite{xiong2019open}, StackExchange \cite{yuan-etal-2020-one} and KPTimes \cite{gallina2019kptimes}. The title and abstract of an article are concatenated to form a testing document.


The testing datasets are categorized into \textit{in-domain} and \textit{out-of-domain}, by measuring the percentage of keyphrase overlap with the training corpus, i.e. the percentage of golden keyphrases in the testing dataset that also appear in some documents in KP20K. We choose the mean value of $\sim 33$ as a threshold to classify the testing datasets. As a result, the in-domain datasets include SemEval, Inspec, NUS, Krapivin and StackExchange, while the other three are out-of-domain.

In the testing phase, besides using KP20K, we also use the training set of StackExchange (300k documents) and KPTimes (260k documents) to further index the phrase bank and the retriever. The purpose of adding these additional sources in the testing phase is to test whether or not our model can easily integrate additional information to work in domains unseen during training, without having it re-trained.





\begin{table}[]
\resizebox{\columnwidth}{!}{%
\begin{tabular}{|lllllllll|}
\hline
\multicolumn{9}{|c|}{\textbf{Present keyphrase generation}} \\ \hline
\multicolumn{1}{|l|}{\multirow{2}{*}{}} &
  \multicolumn{2}{c|}{DUC-2001} &
  \multicolumn{2}{c|}{KPTimes} &
  \multicolumn{2}{c|}{OpenKP} &
  \multicolumn{2}{c|}{Average} \\ \cline{2-9} 
\multicolumn{1}{|l|}{} &
  F1@3 &
  \multicolumn{1}{l|}{F1@5} &
  F1@3 &
  \multicolumn{1}{l|}{F1@5} &
  F1@3 &
  \multicolumn{1}{l|}{F1@5} &
  F1@3 &
  F1@5 \\ \hline
\multicolumn{1}{|l|}{TF-IDF} &
  8.4 &
  \multicolumn{1}{l|}{11.3} &
  \textbf{21.2} &
  \multicolumn{1}{l|}{\textbf{21.5}} &
  {\ul 13.5} &
  \multicolumn{1}{l|}{{\ul 12.9}} &
  14.4 &
  15.2 \\
\multicolumn{1}{|l|}{TextRank} &
  9.8 &
  \multicolumn{1}{l|}{14} &
  10.1 &
  \multicolumn{1}{l|}{9.7} &
  9.8 &
  \multicolumn{1}{l|}{9.3} &
  9.9 &
  11 \\
\multicolumn{1}{|l|}{MultipartiteRank} &
  13.7 &
  \multicolumn{1}{l|}{18.7} &
  18 &
  \multicolumn{1}{l|}{18.7} &
  12.7 &
  \multicolumn{1}{l|}{11.9} &
  {\ul 14.8} &
  {\ul 16.4} \\
\multicolumn{1}{|l|}{EmbedRank} &
  \textbf{20} &
  \multicolumn{1}{l|}{\textbf{24.5}} &
  10 &
  \multicolumn{1}{l|}{11.8} &
  6.9 &
  \multicolumn{1}{l|}{7.3} &
  12.3 &
  14.5 \\
\multicolumn{1}{|l|}{Global-Local Rank} &
  14.6 &
  \multicolumn{1}{l|}{{\ul 21.8}} &
  10.3 &
  \multicolumn{1}{l|}{10.7} &
  11.7 &
  \multicolumn{1}{l|}{10.1} &
  12.2 &
  14.2 \\ \hline
\multicolumn{1}{|l|}{AutoKeyGen} &
  7.4$_6$ &
  \multicolumn{1}{l|}{9.9$_6$} &
  15.9$_4$ &
  \multicolumn{1}{l|}{16.8$_3$} &
  8.5$_3$ &
  \multicolumn{1}{l|}{8.9$_3$} &
  10.6$_3$ &
  11.8$_3$ \\
\multicolumn{1}{|l|}{\textbf{Ours}} &
  {\ul 15.2$_3$} &
  \multicolumn{1}{l|}{18.2$_4$} &
  {\ul 20.1$_1$} &
  \multicolumn{1}{l|}{{\ul 21.1$_1$}} &
  \textbf{15.9$_8$} &
  \multicolumn{1}{l|}{\textbf{14.2$_2$}} &
  \textbf{17$_3$} &
  \textbf{17.8$_2$} \\ \hline
\multicolumn{1}{|l|}{Supervised - CopyRNN} &
  6.9$_4$ &
  \multicolumn{1}{l|}{8.5$_2$} &
  19.6$_7$ &
  \multicolumn{1}{l|}{19.6$_5$} &
  10.4$_4$ &
  \multicolumn{1}{l|}{10.1$_3$} &
  12.3$_4$ &
  12.7$_3$ \\ \hline
\multicolumn{9}{|c|}{\textbf{Absent keyphrase generation}} \\ \hline
\multicolumn{1}{|l|}{\multirow{2}{*}{}} &
  \multicolumn{2}{c|}{DUC-2001} &
  \multicolumn{2}{c|}{KPTimes} &
  \multicolumn{2}{c|}{OpenKP} &
  \multicolumn{2}{l|}{\multirow{6}{*}{}} \\ \cline{2-7}
\multicolumn{1}{|l|}{} &
  R@5 &
  \multicolumn{1}{l|}{R@10} &
  R@5 &
  \multicolumn{1}{l|}{R@10} &
  R@5 &
  \multicolumn{1}{l|}{R@10} &
  \multicolumn{2}{l|}{} \\ \cline{1-7}
\multicolumn{1}{|l|}{UKE methods} &
  0 &
  \multicolumn{1}{l|}{0} &
  0 &
  \multicolumn{1}{l|}{0} &
  0 &
  \multicolumn{1}{l|}{0} &
  \multicolumn{2}{l|}{} \\ \cline{1-7}
\multicolumn{1}{|l|}{AutoKeyGen} &
  - &
  \multicolumn{1}{l|}{-} &
  0.2$_0$ &
  \multicolumn{1}{l|}{0.3$_0$} &
  - &
  \multicolumn{1}{l|}{-} &
  \multicolumn{2}{l|}{} \\
\multicolumn{1}{|l|}{\textbf{Ours}} &
  - &
  \multicolumn{1}{l|}{-} &
  \textbf{3$_1$} &
  \multicolumn{1}{l|}{\textbf{3.6$_0$}} &
  - &
  \multicolumn{1}{l|}{-} &
  \multicolumn{2}{l|}{} \\ \cline{1-7}
\multicolumn{1}{|l|}{Supervised - CopyRNN} &
  - &
  \multicolumn{1}{l|}{-} &
  0.2$_0$ &
  \multicolumn{1}{l|}{0.4$_1$} &
  - &
  \multicolumn{1}{l|}{-} &
  \multicolumn{2}{l|}{} \\ \hline
\end{tabular}%
}
\caption{Results on out-of-domain datasets.}
\label{tab:out_of_domain_performance}
\end{table}


\subsection{Baselines \& evaluation metrics}
\textbf{Baselines.} We adopt five unsupervised keyphrase extraction (UKE) algorithms, namely TF-IDF, TextRank\footnote{\url{https://github.com/boudinfl/pke}} \cite{mihalcea-tarau-2004-textrank}, MultiPartiteRank\footnote{See footnote 1} \cite{boudin-2018-unsupervised}, EmbedRank \cite{bennani-smires-etal-2018-simple} and Global-Local Rank\footnote{\url{https://github.com/xnliang98/uke_ccrank}} \cite{liang-etal-2021-unsupervised} as baselines.

We also compare our model with AutoKeyGen \cite{Shen_Wang_Meng_Shang_2022}, which is the only previous work on unsupervised keyphrase generation. With the permission from the authors, we implemented and report the AutoKeyGen-Copy version. Furthermore, we present CopyRNN \cite{meng-etal-2017-deep} as a supervised baseline. We employ the Transformer-based pointer-generator network for both AutoKeyGen and CopyRNN, with the same settings as described in \ref{sec:phraseness_module_implementation}. Both AutoKeyGen and CopyRNN are trained using KP20K.

\textbf{Evaluation metrics.} We follow the widely-used strategy and separate the evaluation of present and absent keyphrase generation. We employ macro-average F1 and macro-average Recall for evaluating present and absent keyphrase generation, respectively. We evaluate present keyphrases at top 3 and 5 predictions; and absent keyphrases at top 5 and 10. The predictions as well as the groundtruths are stemmed using Porter Stemmer\footnote{\url{https://github.com/nltk/nltk/blob/develop/nltk/stem/porter.py}} \cite{porter1980algorithm} and duplicates are removed before evaluation.

\subsection{Results}

\begin{figure*}[h]
\begin{center}
\includegraphics[width=14cm]{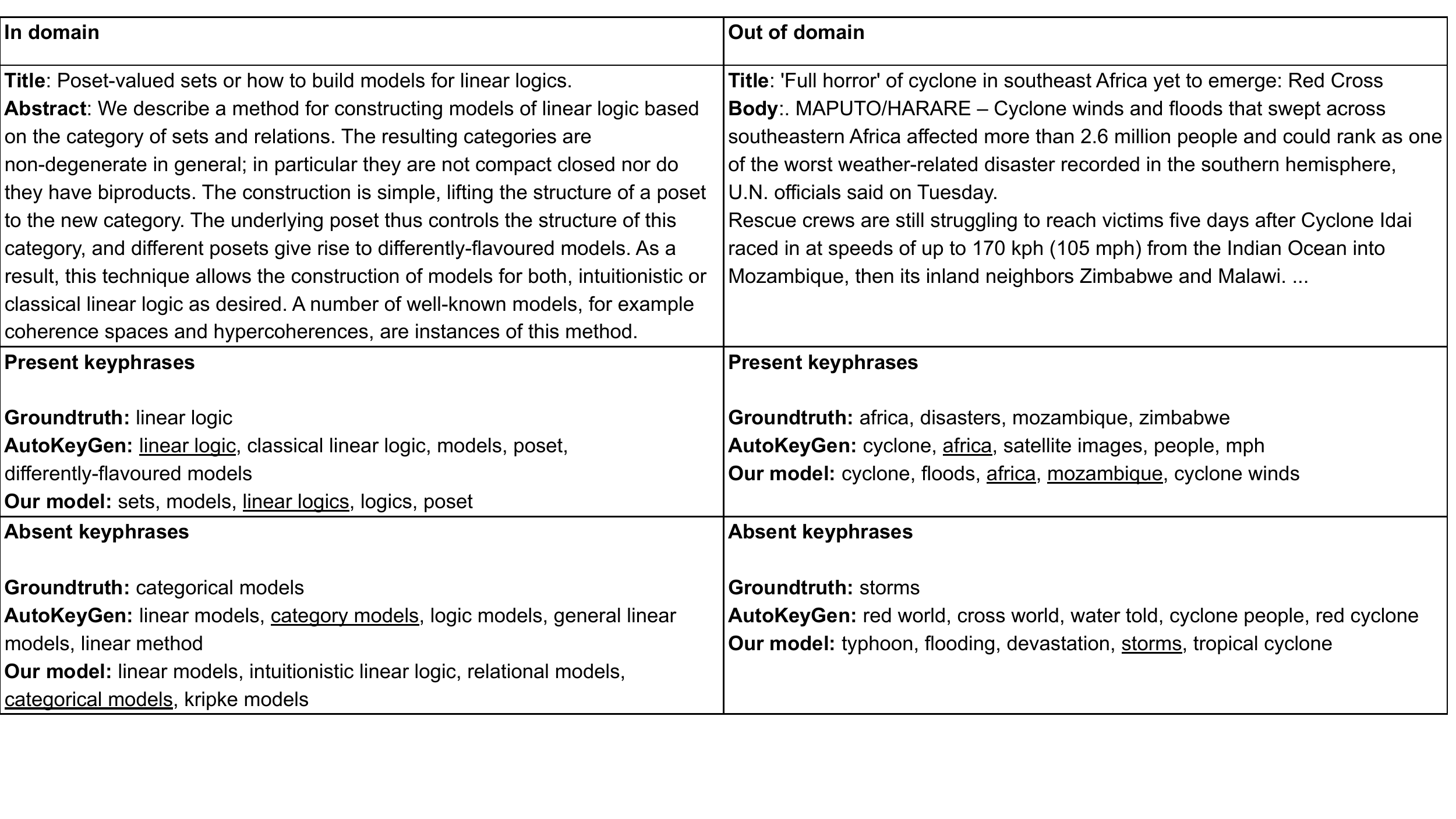}
\end{center}
\caption{Two examples of the generated keyphrases from AutoKeyGen and our proposed model. We illustrate the top 5 predictions. Correctly predicted keyphrases are underlined.}
\label{fig:casestudy}
\end{figure*}

\subsubsection{Keyphrase generation for in-domain cases}

Table \ref{tab:in_domain_performance} illustrates the performance of our proposed model and the baselines for the five in-domain datasets. We also display the average performance across datasets.

\textbf{Present keyphrase generation.} For predicting present keyphrases, our model is best or second-best on most datasets. On SemEval, our model is slightly inferior to TF-IDF and Global-Local Rank. The results on Inspec are worth noting, as our proposed model is significantly outperformed by UKE methods. This inferior performance may be due to this dataset not favoring generative methods, as even CopyRNN, the supervised baseline, failed to compete with UKE methods on Inspec. This behavior has also been observed in a recent work \cite{gallina2020large}. Although not being able to outperform existing methods on all datasets, our proposed model still achieves the best weighted-average results, outperforming the second-best by about 14\% for top 3 predictions and 10\% for top 5.

\textbf{Absent keyphrase generation.} For predicting absent keyphrases, our proposed model outperforms existing work on all datasets. UKE methods cannot be compared with our model, as they only extract present keyphrases. When comparing with AutoKeyGen, we observe that our proposed model have significantly better performance, except for the Inspec dataset where the results are on par. On average, we outperform AutoKeyGen by nearly twice for both top 5 and top 10 predictions.

\subsubsection{Keyphrase generation for out-of-domain cases}

One important objective of this work is the proposed model's capability to perform in out-of-domain settings. We show present and absent keyphrase generation performance for out-of-domain datasets in table \ref{tab:out_of_domain_performance}. For absent keyphrase generation, we only report results on KPTimes, as DUC-2001 and OpenKP mainly contain present keyphrases.

\textbf{Present keyphrase generation.} Our model achieves the best or second-best results on all out-of-domain datasets. Similar to the in-domain cases, our model achieves the best weighted-average results despite not being able to outperform all baselines on all datasets. Of the two unsupervised keyphrase generation methods, our proposed model achieves significantly better results than AutoKeyGen in the out-of-domain setting. 

\textbf{Absent keyphrase generation. } In the out-of-domain setting. It can be seen that AutoKeyGen fails to generate absent keyphrases, with the recall of only $0.3\%$ for top 10 predictions. On the other hand, our model can recall 3.6\% of absent keyphrases. This improvement is significant considering that absent keyphrase generation has been pointed out to be a ``very challenging task'' \cite{meng-etal-2017-deep}.

\subsubsection{Comparison to supervised baseline}

Although not being able to compete with the supervised baseline on the in-domain datasets, our model has narrowed the gap between supervised and unsupervised keyphrase generation methods. In addition, our model shows remarkable performance on out-of-domain datasets, while the supervised baseline shows poor generalization. It can be seen from Table \ref{tab:in_domain_performance} and \ref{tab:out_of_domain_performance} that the performance of the supervised baseline plummets on out-of-domain datasets. On the other hand, our model is able to retain performance across domains.

\subsection{Ablation study}

We perform an ablation study to further understand the role of the components of our proposed model. In particular, we test our model with some components removed, namely the adjustment weight defined in Equation \ref{eq:adjustment_weight}, the references and the part-of-speech information. We report the results in Table \ref{tab:ablation_study}. For KPTimes and OpenKP, we sample 200 and 500 documents from their original validation and test set to perform the ablation study.

We observe that no model in the ablation study achieve the best performance in all cases. However, the full version of our model shows to be more well-rounded compared to its ablations. 

\textbf{Firstly}, the adjustment weight $b(\boldsymbol{y})$ proves to be crucial, as removing it cause our model's performance to drop in most cases. This confirms that the positional information is useful in predicting present keyphrases, as has been pointed out by previous work \cite{florescu-caragea-2017-positionrank, gallina2020large}. Moreover, prioritizing phrases that exist in the phrase bank also proves to be effective for predicting absent keyphrases. \textbf{Next}, removing the references shows to heavily affect absent keyphrase generation, especially on the out-of-domain dataset KPTimes. On the other hand, present keyphrase generation seems not to be affected without using references. \textbf{Finally}, the version of our model without part-of-speech information is able to maintain present keyphrase generation performance for the in-domain dataset (Krapivin), but slightly worsens when being moved out-of-domain. For absent keyphrase generation, it seems that part-of-speech information does not help for KPTimes. A possible explanation is that KPTimes mostly contains single-word keyphrases and therefore grammatical information can offer little help in this case.

\begin{table}[]
\resizebox{\columnwidth}{!}{%
\begin{tabular}{|lllll|}
\hline
\multicolumn{5}{|c|}{\textbf{Present keyphrase generation}}                                                                                                          \\ \hline
\multicolumn{1}{|l|}{\multirow{2}{*}{}}    & \multicolumn{1}{l|}{Krapivin}      & \multicolumn{1}{l|}{DUC-2001}      & \multicolumn{1}{l|}{KPTimes}       & OpenKP   \\ \cline{2-5} 
\multicolumn{1}{|l|}{}                     & \multicolumn{1}{l|}{F1@5}          & \multicolumn{1}{l|}{F1@5}          & \multicolumn{1}{l|}{F1@5}          & F1@5     \\ \hline
\multicolumn{1}{|l|}{No adjustment weight} & \multicolumn{1}{l|}{17.4}          & \multicolumn{1}{l|}{\textbf{19.3}} & \multicolumn{1}{l|}{21.3}          & 10       \\
\multicolumn{1}{|l|}{No references} & \multicolumn{1}{l|}{{\ul 20.7}} & \multicolumn{1}{l|}{{\ul 18.8}} & \multicolumn{1}{l|}{21.4} & \textbf{14.6} \\
\multicolumn{1}{|l|}{No POS}               & \multicolumn{1}{l|}{\textbf{21.1}} & \multicolumn{1}{l|}{17.6}          & \multicolumn{1}{l|}{21.5}          & 13.1     \\
\multicolumn{1}{|l|}{Full}                 & \multicolumn{1}{l|}{20.5}          & \multicolumn{1}{l|}{18.2}          & \multicolumn{1}{l|}{\textbf{21.8}} & {\ul 14} \\ \hline
\multicolumn{5}{|c|}{\textbf{Absent keyphrase generation}}                                                                                                           \\ \hline
\multicolumn{1}{|l|}{\multirow{2}{*}{}}    & \multicolumn{1}{l|}{Krapivin}      & \multicolumn{1}{l|}{DUC-2001}      & \multicolumn{1}{l|}{KPTimes}       & OpenKP   \\ \cline{2-5} 
\multicolumn{1}{|l|}{}                     & \multicolumn{1}{l|}{R@10}          & \multicolumn{1}{l|}{R@10}          & \multicolumn{1}{l|}{R@10}          & R@10     \\ \hline
\multicolumn{1}{|l|}{No adjustment weight} & \multicolumn{1}{l|}{5.8}           & \multicolumn{1}{l|}{-}             & \multicolumn{1}{l|}{2.5}           & -        \\
\multicolumn{1}{|l|}{No references}        & \multicolumn{1}{l|}{5.5}           & \multicolumn{1}{l|}{-}             & \multicolumn{1}{l|}{1}             & -        \\
\multicolumn{1}{|l|}{No POS}               & \multicolumn{1}{l|}{{\ul 7.1}}     & \multicolumn{1}{l|}{-}             & \multicolumn{1}{l|}{\textbf{3.5}}  & -        \\
\multicolumn{1}{|l|}{Full}                 & \multicolumn{1}{l|}{\textbf{7.2}}  & \multicolumn{1}{l|}{-}             & \multicolumn{1}{l|}{{\ul 3.2}}     & -        \\ \hline
\end{tabular}%
}
\caption{Ablation study.}
\label{tab:ablation_study}
\end{table}

\section{Case study}

We display two examples of generated keyphrases from AutoKeyGen and our proposed model in Figure \ref{fig:casestudy}. The first example is from Krapivin, an in-domain dataset, while the second one is from KPTimes, an out-of-domain dataset. For the first example, we observe that both the proposed model and AutoKeyGen correctly predict the groundtruth (present and absent) keyphrases. However, it can be seen that, for generating absent keyphrases, AutoKeyGen only reorders words that are present in the given text. On the other hand, our model can generate keyphrases whose component words are absent, such as ``relational models'', ``categorical models'' and ``kripke models''.

In the second example, it is clear that our model predicts more correct keyphrases. We observe that the absent keyphrases generated by AutoKeyGen are highly irrelevant. On the other hand, our model successfully predicts ``storms'' and also outputs other absent keyphrases that are relevant, although not being within the ground truth keyphrases. This example help shows that our model is better at handling documents from different domains.

\section{Related work}
\label{sec:related_word}
\subsection{Unsupervised keyphrase extraction}
Unsupervised keyphrase extraction (UKE) aims at identifying keyphrases within the text. Currently, there are three main classes of UKE methods, namely statistics-based, graph-based and embedding-based. Statistics-based methods \cite{campos2018yake} employ features such as TF-IDF, word position and casing aspect, to determine the relevance of a candidate phrase. 

Graph-based methods typically build a graph from the source text, where a node could be a word or a phrase. Then, different graph-theoretic measures are used to estimate the importance of nodes, and finally phrases are formed based on the top ranked nodes. TextRank \cite{mihalcea-tarau-2004-textrank} builds a word graph where a link between two words exists if they co-occur within a window. SingleRank \cite{wan2008single}, CiteTextRank \cite{gollapalli2014extracting} employs related documents to better measure similarity between word nodes. TopicRank \cite{bougouin2013topicrank}, Topical PageRank \cite{liu2010automatic} incorporate topical information in the graph ranking algorithm. Positional information is used in PositionRank \cite{florescu2017positionrank} to favor keyphrases that appear earlier in the text. \cite{boudin-2018-unsupervised} utilizes the structure of multi-partite graphs to extract diverse keyphrases.

Embedding-based methods utilize embedding spaces to measure informativeness of candidates. EmbedRank \cite{bennani-smires-etal-2018-simple} rank candidates by measuring their distance to the source text in a pretrained sentence embedding space, then an optional diversification step is performed using maximal-marginal relevance to ensure diversity of extracted keyphrases. \cite{liang-etal-2021-unsupervised} jointly models local and global context of the document when ranking candidates.

\subsection{Unsupervised keyphrase generation}
Keyphrase generation aims at predicting both present and absent keyphrases for the source text. To our best knowledge, AutoKeyGen \cite{Shen_Wang_Meng_Shang_2022} is currently the only unsupervised keyphrase generation method. AutoKeyGen trains a seq2seq model on automatically generated silver labeled document-keyphrase pairs. The silver keyphrases are both present and absent, where present ones are extracted, and the absent ones are constructed from the words present in the text.

\section{Conclusions}
In this paper, we propose a new problem called unsupervised open-domain keyphrase generation. We propose a seq2seq model that consists of two modules, one is responsible for generating phrases while the other guides the generation towards phrases that reflect the core concepts of the given text. Our experiments on eight benchmark datasets from multiple domains demonstrate that our model outperforms existing unsupervised methods and narrows the gap between unsupervised and supervised keyphrase generation models. Furthermore, we demonstrate that the proposed model can perform consistently across domains.

\section*{Limitations}
One limitation of the proposed method is that it does not consider domain-specific information to evaluate informativeness. The phraseness module has access to domain-specific knowledge, which are the phrases that occur in similar contexts, i.e. the references. On the other hand, the informativeness module only employs a domain-general sentence embedding model to measure informativeness of phrases. Therefore, the integration of both domain-specific and domain-general information for the evaluation of informativeness may be worth further investigation. 

Another limitation of this work is that we only tested the proposed method on short texts. Therefore, it is uncertain of the proposed framework’s performance on
long text documents. Handling long texts could be significantly more difficult than
short text, as long texts contain much more information (can discuss a variety of
topics).

The final limitation of this work is the absence of experiments on using different sentence embedding models to construct the informativeness module. Therefore, it might be useful to explore the impact of different sentence embedding models on keyphrase generation performance. We leave this for future work.

\bibliography{anthology,custom}
\bibliographystyle{acl_natbib}

\appendix

\section{Implementation details}
\label{sec:implementation_details}

\subsection{Phraseness module}
\label{sec:phraseness_module_implementation}

\textbf{Retriever. } We extract noun phrases for the documents in the training set of KP20K, StackExchange and KPTimes; and form the phrase bank by keeping the noun phrases that occur in at least 5 documents. For obtaining embeddings of documents, we employ the Sent2Vec pretrained model named {\fontfamily{qcr}\selectfont sent2vec\_wiki\_unigrams}\footnote{\url{https://github.com/epfml/sent2vec}}, a 600-dimensional sentence embedding model trained on English Wikipedia. We utilize Faiss\footnote{\url{https://github.com/facebookresearch/faiss}} \cite{johnson2019billion} for indexing the phrases and their context embeddings.

\textbf{Seq2seq model.} Both the encoder and decoder of the seq2seq model contains 3 layers. The model dimensionality and the word embedding size are both set to 256, and the part-of-speech embedding size is set to 64. The seq2seq model employs attention with 8 heads. The encoder and decoder have separate vocabularies, both contain 40000 words. The encoder and decoder vocabulary contains the frequent words in the unlabeled corpus and among the extracted noun phrases, respectively. The seq2seq model is optimized using Adam optimizer \cite{kingma2014adam}, with a learning rate of 0.0001, gradient clipping = 0.1 and a dropout rate of 0.1. We trained our model in 15 epochs. After every 3 training epoch, the learning rate is reduced by $10\%$. The seq2seq model contains 34M trainable parameters, and training it for 15 epochs took about 7 hours on a single NVIDIA A40 GPU. We roughly estimate that conducting our experiments, which include training the baseline models, took a total of 150 GPU hours.

\subsection{Informativeness module}
For the informativeness module, we also employ the pretrained Sent2vec model {\fontfamily{qcr}\selectfont sent2vec\_wiki\_unigrams}, to obtain embeddings of words and texts.

\subsection{Keyphrase decoding}
We employ beam search with beam size = 100 and beam depth = 6. The balancing hyperparameter $\lambda$ is set to 0.75. Considering the adjustment weight in Equation \ref{eq:adjustment_weight}, we set $\beta=5/6$ to favor existing phrases in the phrase bank $B$. For each text, we retrieve 15 references, some of which can be filtered out by the threshold $\tau$, which is set to 0.7.

We use the validation set of each testing dataset to select the value for the length penalty factor $\alpha$. In particular, we choose $\alpha \in$ $\{$-1, -0.75, -0.5, -0.25, 0, 0.25, 0.5, 0.75, 1$\}$ that maximizes the geometric mean of the evaluation metrics, i.e. F1 at top 3 and 5 for present keyphrases and Recall at top 5 and 10 for absent keyphrases. Since the value range of these metrics are different from one another, we divide each metric by its maximum value (that can be found as we try different values of $\alpha$) for normalization before taking the geometric mean. Table \ref{tab:length_penalty} provide the best $\alpha$ values for each dataset in each run in our experiment.

\begin{table}[]
\resizebox{\columnwidth}{!}{%
\begin{tabular}{|l|c|c|c|c|c|}
\hline
              & 1st run & 2nd run & 3rd run & 4th run & 5th run \\ \hline
SemEval       & -1      & -1      & -0.5    & -1      & -1      \\
Inspec        & -1      & -1      & -1      & -0.75   & -1      \\
NUS           & -0.75   & -0.75   & -0.25   & -0.25   & -0.5    \\
Krapivin      & -0.25   & -0.5    & -0.25   & -0.25   & -0.5    \\
StackExchange & 1       & 1       & 1       & 1       & 1       \\
DUC-2001      & -1      & -1      & -1      & -1      & -1      \\
KPTimes       & 0.5     & 0.5     & 0.75    & 0.5     & 1       \\
OpenKP        & 0.5     & -0.25   & 0.25    & -0.25   & -0.25   \\ \hline
\end{tabular}%
}
\caption{Best length penalty values for each dataset in each run.}
\label{tab:length_penalty}
\end{table}

\end{document}